\documentclass[letterpaper, 10 pt, conference]{ieeeconf}  
\usepackage{bm}
\usepackage{dcolumn}
\usepackage{latexsym}
\usepackage{dcolumn}
\usepackage{epsf}
\usepackage{hyperref}
\usepackage{graphicx} 
\usepackage{epsfig}
\usepackage{epstopdf}
\usepackage{tikz}
\usepackage{mathptmx} 
\usepackage{times} 
\usepackage{amsmath} % assumes amsmath package installed
\usepackage{amssymb}  % assumes amsmath package installed
\usepackage{cite}
\usepackage{graphicx}
\usepackage{tabularx}
\usepackage{subcaption}
\usepackage{caption}
\usepackage{float}
\usepackage{gensymb}
\usepackage{color}
\usepackage{soul}

\IEEEoverridecommandlockouts                              
\title{ \LARGE \bf Dynamic Flex-and-Flip Manipulation of Deformable Linear Objects}

\author{Chunli Jiang, Abdullah Nazir, Ghasem Abbasnejad, and Jungwon Seo
\thanks{All the authors are with The Hong Kong University of Science and Technology, Clear Water Bay, Hong Kong (Chunli Jiang and Ghasem Abbasnejad: Department of Mechanical and Aerospace Engineering {\tt\small cjiangab@connect.ust.hk, abbasnejad@ust.hk}; Abdullah Nazir: Department of Electronic and Computer Engineering {\tt\small sanazir@connect.ust.hk}; Jungwon Seo: Departments of Mechanical and Aerospace/-Electronic and Computer Engineering {\tt\small junseo@ust.hk}).}%
}

\begin{document}

\maketitle
\thispagestyle{empty}
\pagestyle{empty}

%%%%%%%%%%%%%%%%%%%%%%%%%%%%%%%%%%%%%%%%%%%%%%%%%%%%%%%%%%%%%%%%%%% ABSTRACT
\begin{abstract}

This paper presents the technique of \textit{flex-and-flip} manipulation. It is suitable for grasping thin, flexible linear objects lying on a flat surface. During the manipulation process, the object is first flexed by a robotic gripper whose fingers are placed on                                                                                                                                                                                                                                                                                                          top of it, and later the increased internal energy of the object helps the gripper obtain a stable pinch grasp while the object flips into the space between the fingers. The dynamic interaction between the flexible object and the gripper is elaborated by analyzing how energy is exchanged. We also discuss the condition on friction to prevent loss of contact. Our flex-and-flip manipulation technique can be implemented with open-loop control and lends itself to underactuated, compliant finger mechanism. A set of experiments in robotic page turning performed with our customized hardware and software system demonstrates the effectiveness and robustness of the manipulation technique.

\end{abstract}

%%%%%%%%%%%%%%%%%%%%%%%%%%%%%%%%%%%%%%%%%%%%%%%%%%%%%%%%%%%%%%%%%%% INTRODUCTION
\section{Introduction}

%%%%%%%%%%%%Figure:Steps of page turning
\begin{figure}[!htb]
\centering
\begin{subfigure}[b]{8.1cm} 
\includegraphics[width=8.1cm]{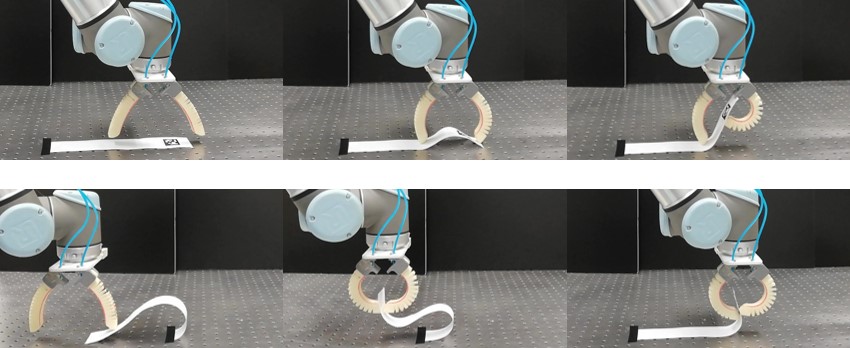}
\end{subfigure}

\caption{Sequence of snapshots (clockwise from the top left panel) showing our flex-and-flip manipulation technique applied to the task of page turning with a two-fingered soft robotic hand.
}
\label{fig:technique_demo}
\vspace{-0.2cm}
\end{figure}

This paper investigates a robotic manipulation technique that we call {\it flex-and-flip} manipulation. The technique is targeted at flexible, thin objects and applied to obtaining secure, pinch grasps on them. Fig.~\ref{fig:technique_demo} shows the progress of the operation that is applied to the task of page turning, with the two-fingered soft robotic hand. Initially, the flexible object (a strip of paper) is placed on a tabletop. The robotic hand approaches and deforms the object (the ``flex'' phase of the operation), and then the deformed object is tucked into the gap between the fingers (the ``flip'' phase). Finally, a pinch grasp is obtained and the page turning task (or other secondary manipulation tasks) can be performed.

The task of grasping flexible objects seems to be everywhere, in a wide range of industry sectors and many conceivable service robotics scenarios. However, the complexity of flexible objects both in terms of geometry change due to deformation and intrinsic properties poses a great challenge, along with the well-known difficulties in rigid body manipulation such as friction modeling. Our flex-and-flip manipulation technique takes advantage of the dynamics of deformed flexible objects to facilitate grasp acquisition by helping the object flip and separate the two fingers in a timely manner. The soundness of the approach will be discussed by analyzing how the internal energy of the object-robot system changes over time.

As will be demonstrated by our experiments, the technique of flex-and-flip manipulation can be implemented with a minimalist hardware setting featuring 3D printed soft robotic fingers operated in an underactuated manner, which will also help negotiate the interaction with the environment, that is, the table top supporting the flexible object. Closed-loop motion/force control is not necessary. Open-loop playback of finger motion can guarantee highly repetitive, successful results. In addition to the main challenge task of this study, page turning demonstrated in Fig.~\ref{fig:technique_demo}, we expect that various secondary manipulation tasks on flexible objects can be facilitated based on the secure grasps obtained by our flex-and-flip manipulation.

After outlining related work in Sec.~\ref{sec:relatedwork} and presenting our problem statement in Sec.~\ref{sec:prob_desc}, our manipulation technique for grasping flexible objects is discussed in Sec.~\ref{sec:theory}. The implementation of the technique is presented in Sec.~\ref{sec:implementation} with a set of experiments.

%%%%%%%%%%%%%%%%%%%%%%%%%%%%%%%%%%%%%%%%%%%%%%%%%%%%%%%%%%%%%%%%%%% RELATED WORK

\section{Related Work}
\label{sec:relatedwork}

Beyond the traditional approach in robotics in which robots composed of rigid links are governed and controlled by rules derived from rigid body mechanics, robots that are soft and/or capable of handling soft objects are becoming of great concern these days. One fundamental problem is how to model the behavior of soft material. In the robotics literature, \cite{Hirai2004IJRR} addresses the static deformation modeling of linear objects based on differential geometry, along with a set of experiments. \cite{Moll_2006} presents a subdivision-based computational approach to deformation modeling, which can be more computationally favorable. Generally, the modeling of object deformation is formulated as an energy minimization problem subject to boundary conditions imposed by the presence of obstacles for example.

Our work here is concerned with object manipulation in the presence of compliance. Knotting/unknotting manipulation of deformable linear objects is one of the problems that have received considerable attention with important potential applications to medical and service robotics: its planning/control aspects are addressed in \cite{Ladd_2004, Wakamatsu_2006, 4359263}; see \cite{Matsuno_2006} for a discussion about perception. \cite{doi:10.1177/0278364911430417} is concerned with the manipulation of two-dimensional deformable objects with an application to laundry folding. Other examples include \cite{Odhner_2013}, where precision grasping of thin objects is attained through flip-and-pinch manipulation with underactuated, compliant finger mechanism. Our challenge task in the present work is page turning. \cite{young2004low} presents a highly reliable page turning technique using mechanical adhesion by polymeric material. Our approach to page turning is to take advantage of nonprehensile, dynamic manipulation (see \cite{lynch1999dynamic, mason2001mechanics} and references therein), which has also been addressed by some of our previous works \cite{8392736, Kim2019Shallow, Nazir_2019_ICRA}.

Soft robotic hands can be effective for handling soft objects. We also adopt a soft hand for the page turning task. The development of underactuated, compliant robotic hands has been an active research topic. An early example can be seen in \cite{hirose_1978}. Recent examples can be seen in \cite{Odhner_2014} featuring an underactuated hand capable of even dexterous manipulation and \cite{Mosadegh_2014} featuring a robotic finger composed of pneumatic chambers capable of fast motion.

%%%%%%%%%%%%%%%%%%%%%%%%%%%%%%%%%%%%%%%%%%%%%%%%%%%%%%%%%%%%%%%%%%% TASK DESCRIPTION
\section{Problem Description}
\label{sec:prob_desc}

The problem we address in the paper concerns a novel robotic manipulation technique that can be applied to the grasping of flexible objects. Compared with rigid objects, grasping flexible objects additionally necessitates proper handling of deformation, which will affect not only object geometry but also contact forces. We are interested in taking advantage of the dynamics of object deformation in the process of grasp acquisition.

The objects of interest in this work are modeled as a {\it deformable linear object} \cite{Hirai2004IJRR} (or simply {\it object} for short), which can be thought of as a flexible wire. We further assume that our objects are elastic yet inextensible. This model can represent a wide range of objects that are practically important, for example, strips of paper or polymer. Initially, an object is supposed to be placed on a flat rigid ground surface, with no deformation. Any deformation happens on the plane containing the object and perpendicular to the surface. We are interested in a manipulation scenario with two point contacts, which can practically be realized with a two-fingered gripper. 

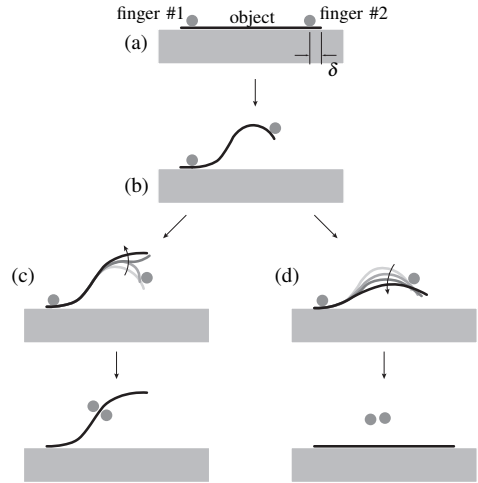
\begin{figure}[h!]
\centering 
\includegraphics[width=2.5in]{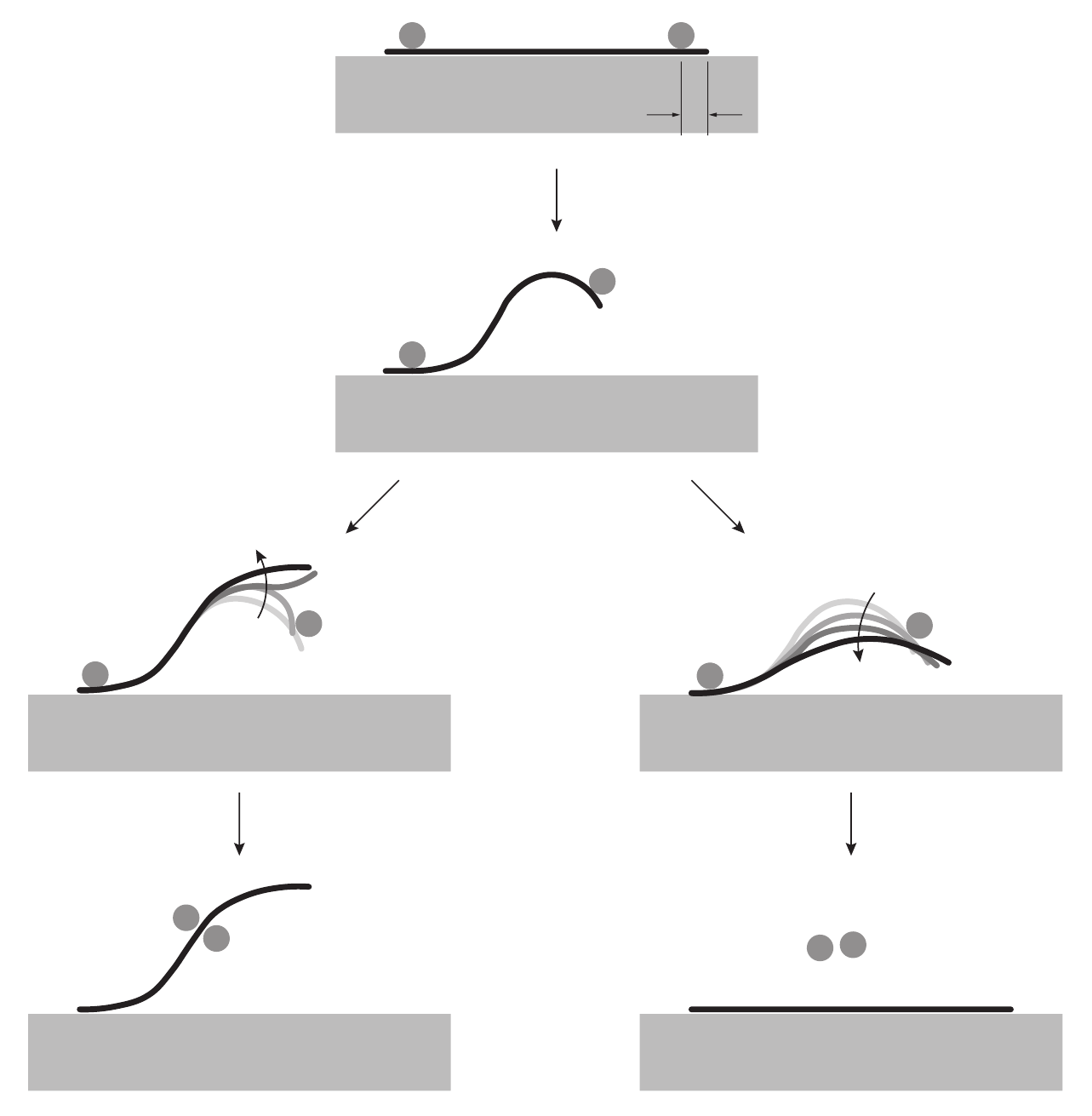}
\setlength{\unitlength}{1cm}
\begin{picture}(0,0)
\put(-5.1,5.9){\footnotesize (a)}
\put(-5.2,6.30){\scriptsize finger \#1}
\put(-2.5,6.30){\scriptsize finger \#2}
\put(-2.4,5.55){\scriptsize $\delta$}
\put(-3.7,6.25){\scriptsize object}
\put(-5.1,4.0){\footnotesize (b)}
\put(-6.6,2.8){\footnotesize (c)}
\put(-3.1,2.8){\footnotesize (d)}
%\put(-3.7,.78){\footnotesize (e)}
%\put(.1,.78){\footnotesize (f)}
\end{picture}
\caption{Progress of our flex-and-flip manipulation with the two point fingers: (a) initial configuration with $\delta$ infinitesimally small, (b) object deformation, (c) successful pinch grasp, and (d) unsuccessful grasping.}
\label{fig:bifurcation} 
\vspace{-0.2cm}
\end{figure}

Fig.~\ref{fig:bifurcation} illustrates one natural course of manipulation for acquiring a pinch grasp on the object with the setting elaborated in the previous paragraph. We call this {\it flex-and-flip} manipulation in the sense that it is involved with flexing the object and the resulting deformation makes the object dynamically flip across the finger (finger \#2 in the figure). The following sections elaborate the dynamic manipulation technique and present an application to the task of page turning.

%%%%%%%%%%%%%%%%%%%%%%%%%%%%%%%%%%%%%%%%%%%%%%%%%%%%%%%%%%%%%%%%%%%% THEORY

\section{Dynamic Flex-and-Flip Manipulation}
\label{sec:theory}

\begin{figure*}[h]
\centering
\begin{subfigure}[b]{0.42\textwidth}
    \centering
    \includegraphics[trim={15cm 2.5cm 12cm 2cm},clip, width=1\textwidth]{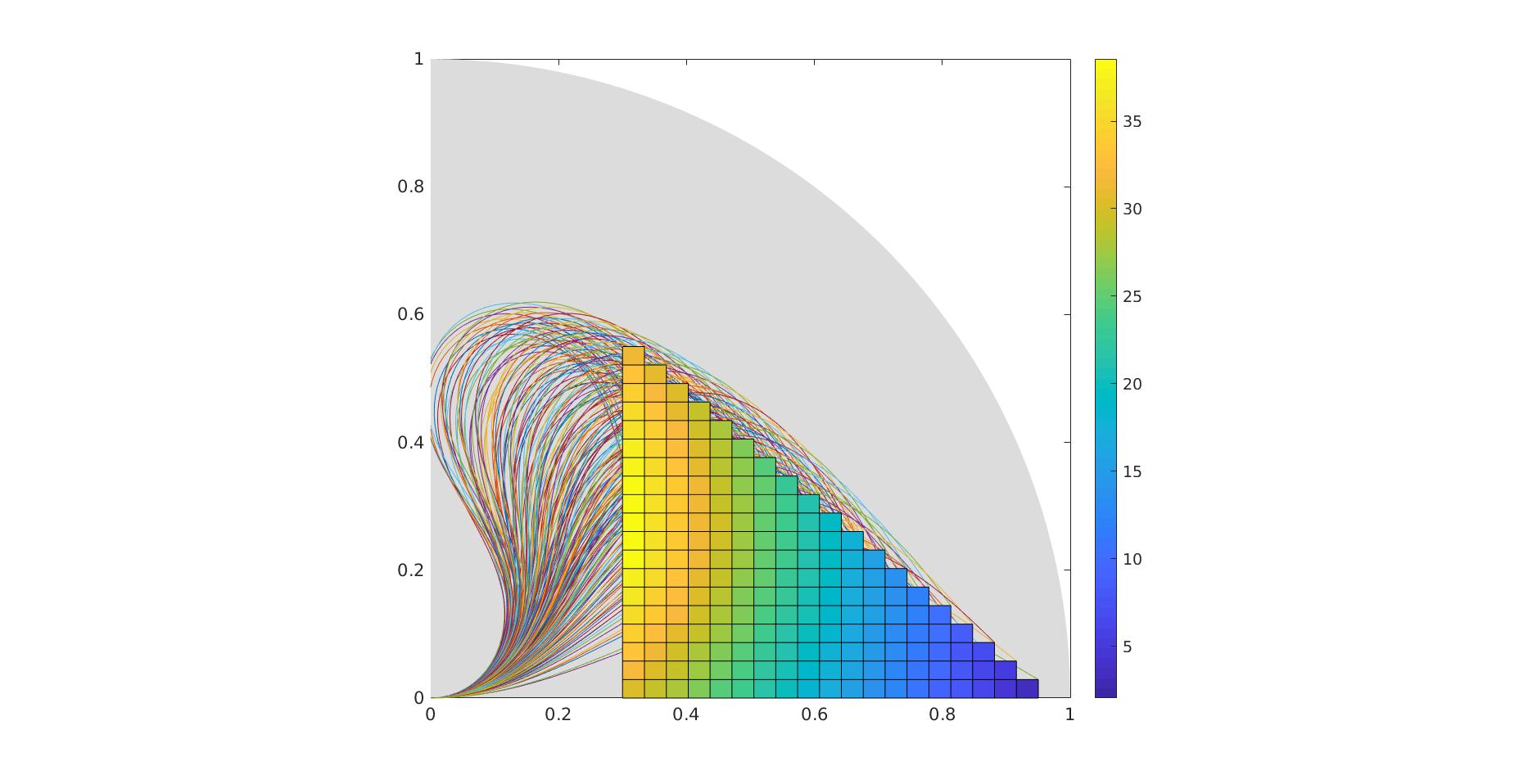}
    \caption{}
    \label{fig:energy_plot}
\end{subfigure}
\begin{subfigure}[b]{0.42\textwidth}
    \centering
    \includegraphics[trim={15cm 2.5cm 12cm 2cm},clip, width=1\textwidth]{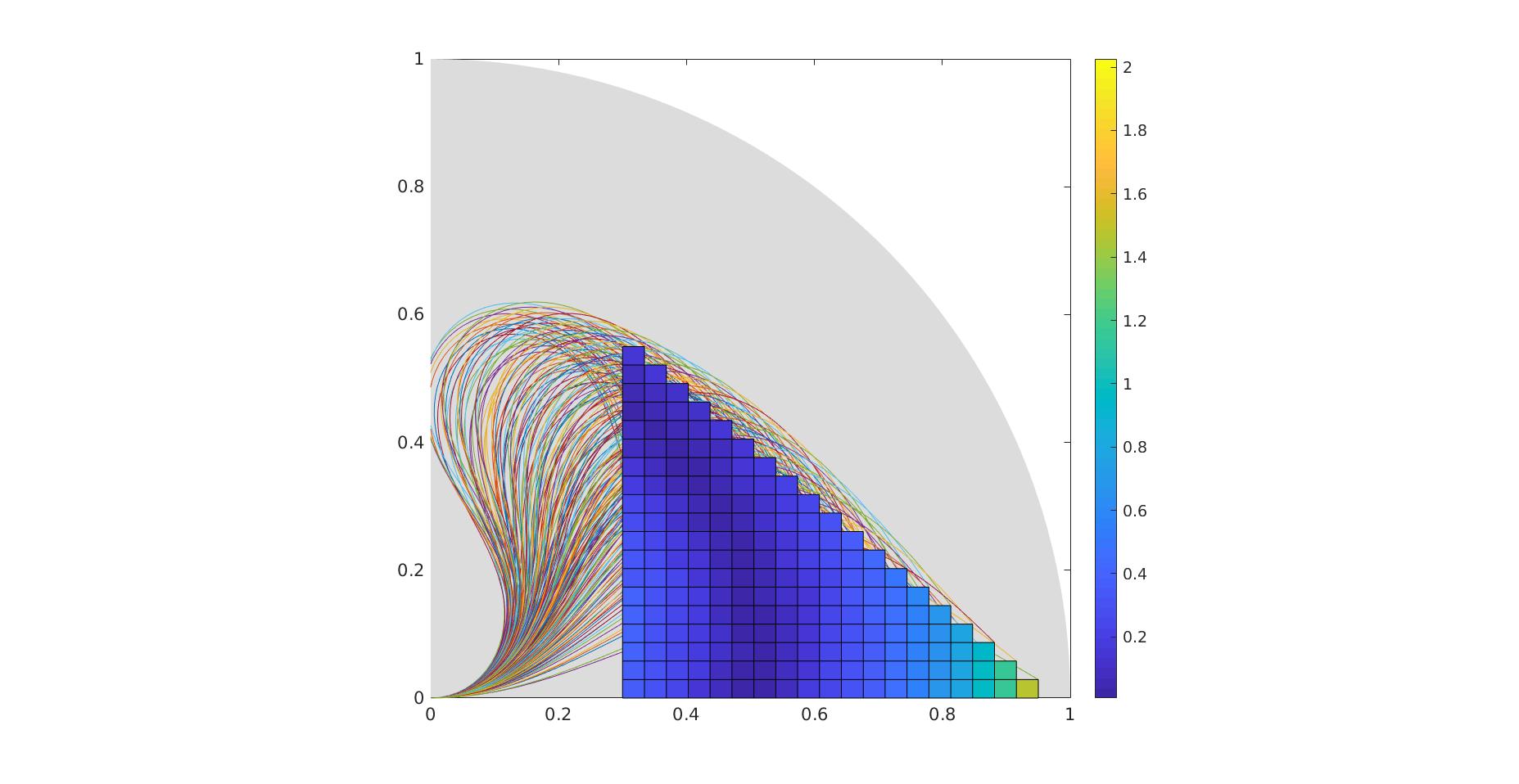}
    \caption{}
    \label{fig:cof}
\end{subfigure}
\setlength{\unitlength}{1cm}
\begin{picture}(0,0)
\put(-8.6,5.8){\footnotesize \rotatebox{270}{Flexural energy (nondimensionalized)}}
\put(-0.9,5.1){\footnotesize \rotatebox{270}{Coefficient of friction}}
\end{picture}
\caption{(a) Computed minimum energy shapes (the S-shaped curves starting from the origin) of a deformable linear object. Each curve ends at one of the grid cells whose color represents the flexural energy value of the shape, as indicated on the vertical color bar, which is nondimensionalized by assuming the flexural rigidity of the object to be 1. The object is deformed on the horizontal axis as in Fig.~\ref{fig:bifurcation}: finger \#1 is located at the origin and the position of finger \#2 varies on the grid. The gray area represents all the possible positions of contact \#2 on the object of length 1. 
(b) Colormap showing the lower bound for the coefficient of friction at contact \#2 to maintain quasistatic equilibrium. All the other settings are the same as (a).
}
\vspace{-0.2cm}
\end{figure*}
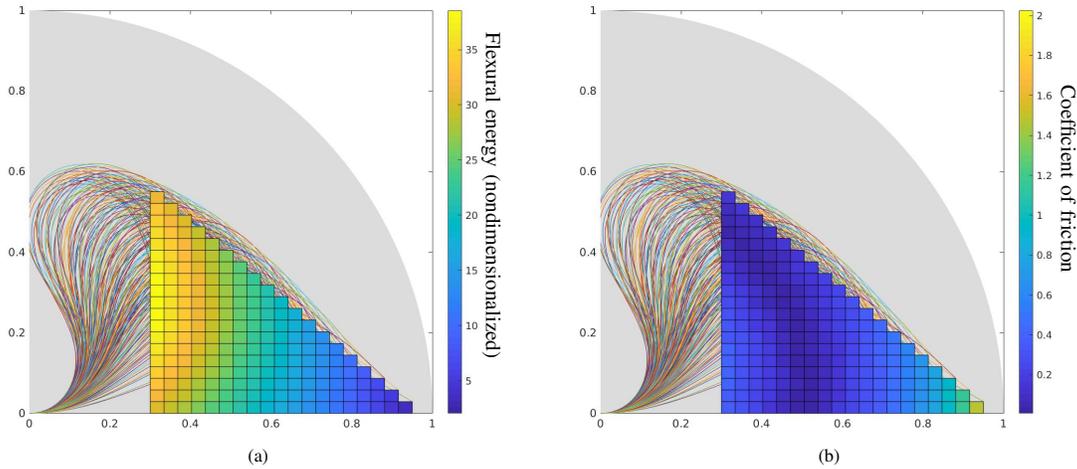

In this section, we present our strategy for successful dynamic flex-and-flip manipulation.

\subsection{Overall Scenario}
\label{subsec:scenario}

As illustrated in Fig.~\ref{fig:bifurcation}, our flex-and-flip manipulation of deformable linear objects begins with the initial configuration in which the two fingers are on the same side of the object (Fig.~\ref{fig:bifurcation}a) and terminates with a pinch grasp in which the fingers are on the different sides (Fig.~\ref{fig:bifurcation}c). Note that $\delta$, the distance between the tip of the object and finger \#2, is assumed to be infinitesimally small. In the meantime, the object is deformed by the fingers (extrinsic factor) and its own energy (intrinsic factor). The progress of the operation will be explained as a two-step process, flex and flip, in the following subsections.

\subsection{Flex: Until Separation of Contact}
\label{subsec:bs}

As the object is deformed from the initial state in which it is lying flat on the ground, its internal energy increases. The internal energy of the object (inextensible, limited to moving on the plane) can be quantified by the flexural energy due to flexural deformation. In order to increase the flexural energy, as can be seen in Fig.~\ref{fig:bifurcation}, the two fingers need to get closer without losing contact with the object. Therefore, it is sensible to press down one of the fingers (finger \#1) on the object and move the other finger (finger \#2) to control flexural deformation, as illustrated in the figure. It is also necessary to lift finger \#2 off the ground; otherwise it can be impossible to finally pinch the object. The object is then expected to deform into an S-shape with a single point of inflection, considering the boundary conditions involved with the course of manipulation: zero (free) slope at contact \#1 with finger \#1 (contact \#2 with finger \#2) and the distance constraint between the fingers. This is confirmed by the simulated object shapes presented in Fig.~\ref{fig:energy_plot}, obtained by finding the curves that minimize the flexural energy \cite{Hirai2004IJRR}: 

\begin{equation}
    U = \frac{1}{2} \int_0^L R_f \kappa^2(s)ds
    \label{eqn:energy_functional}
\end{equation}

(where $s$: the arc length parameter; $L$: the total length, $R_f$: the flexural rigidity, and $\kappa(s)$: the curvature of the object) subject to the boundary conditions.

In a dynamic scenario, finger \#2 (with nonzero inertia) can be controlled to move with positive kinetic energy. As the object is flexed towards the high energy area (the brighter grid cells on Fig.~\ref{fig:energy_plot}), the finger will slow down as its kinetic energy is converted to the flexural energy of the object. Meanwhile, the friction at contact \#2 should be sufficiently large so as to prevent premature loss of contact as illustrated in Fig.~\ref{fig:bifurcation}d. A lower bound for the necessary coefficient of friction can be found by investigating the contact forces at contact \#2: along a nominal trajectory for finger \#2, the contact forces must always lie within their friction cones to prevent loss of contact. The contact forces correspond to the Lagrange multipliers of the energy minimization problem (Eq.~\ref{eqn:energy_functional}). For each energy optimal shape of the object, Fig.~\ref{fig:cof} shows the minimum coefficient of friction at contact \#2 required to maintain static equilibrium. This is obtained by observing the angle between the object shape normal at contact \#2 and the corresponding contact force vector. Note, the lower bound is based on an essentially quasistatic analysis; thus, the actual value might be different in a highly dynamic process.

{\it \textbf{Remark}} Initial contact positioning can be critical to successful flex-and-flip manipulation. In Figs.~\ref{fig:energy_plot} and~\ref{fig:cof}, contact \#2 is assumed to be essentially at the right tip of the object (recall that $\delta$ is assumed to be infinitesimally small in Fig.~\ref{fig:bifurcation}). In practice, $\delta$ needs to be sufficiently large to address errors in sensing and positioning. However, it is sensible to minimize $\delta$ because it can be impossible to flip the object with large $\delta$, considering the actual kinematics of the gripper providing the contacts (this will also be discussed in Sec.~\ref{sec:implementation}).

\subsection{Flip: After Separation of Contact}
\label{subsec:as}

Suppose that the kinetic energy of finger \#2 is set to be less than an intended target value of the flexural energy, due to a safety concern for example. The finger will then eventually come to a stop. At the instant, energy conversion happens in the other direction such that the flexural energy of the object becomes the source of the kinetic energy of the finger. At the same time, the finger loses contact with the object unless the restitution happens in a totally plastic manner. With the loss of contact, the shape of the object will evolve along the negative gradient of the flexural energy field (recall Fig.~\ref{fig:energy_plot}), and this can facilitate the desired flipping motion (Fig.~\ref{fig:bifurcation}c) unless the finger moves totally in sync with the tip of the object. Therefore, in this dynamic setting the internal energy of the object due to its deformation facilitates the flip operation by helping the finger bounce off from the object. We expect that this scenario can be realized in a highly repeatable, robust manner, except for the special case of perfectly plastic collision and zero relative motion between the object and the finger.

\subsection{Discussion}

The dynamic flex-and-flip operation explained in this section does not necessitate closed-loop finger motion/force control. The manipulation technique thus lends itself to underactuated, compliant finger mechanism with open-loop control. Neither tactile sensing nor computation is necessary. The passive mechanics of such mechanism will also help negotiate the interaction between the finger and the ground. These advantages will be confirmed in our experiments to be presented in the next section.

It is also possible to consider an approach based on the traditional position control scheme with high impedance. Finger \#2 can be position controlled to create a gap between the object and the ground surface, possibly in a quasistatic manner with negligible kinetic energy. However, it still seems that fast, dynamic motion is necessary for the finger to get into the gap before it is closed again due to the passive dynamics of the object under gravity. Instead, our approach takes advantage of the dynamics of the flexed object.

%%%%%%%%%%%%%%%%%%%%EXPERIMENTAL VALIDATION
\section{Implementation and Experiments}
\label{sec:implementation}

%%%%%%%%%% Figure:Soft hand configuration 
\begin{figure}[t]
\centering

\begin{subfigure}[t]{0.23\textwidth}
\centering
\includegraphics[width=1\textwidth]{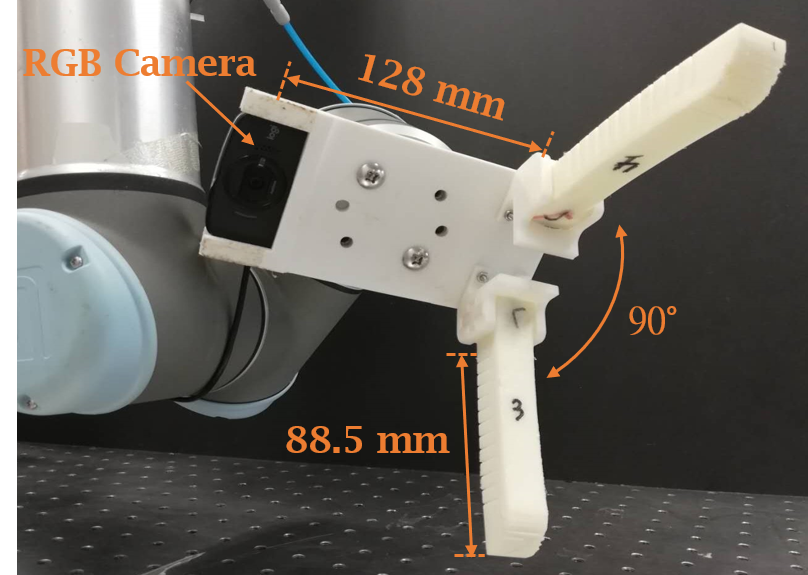}
\caption{}
\end{subfigure}
\begin{subfigure}[t]{0.22\textwidth}
\centering
\includegraphics[width=1\textwidth]{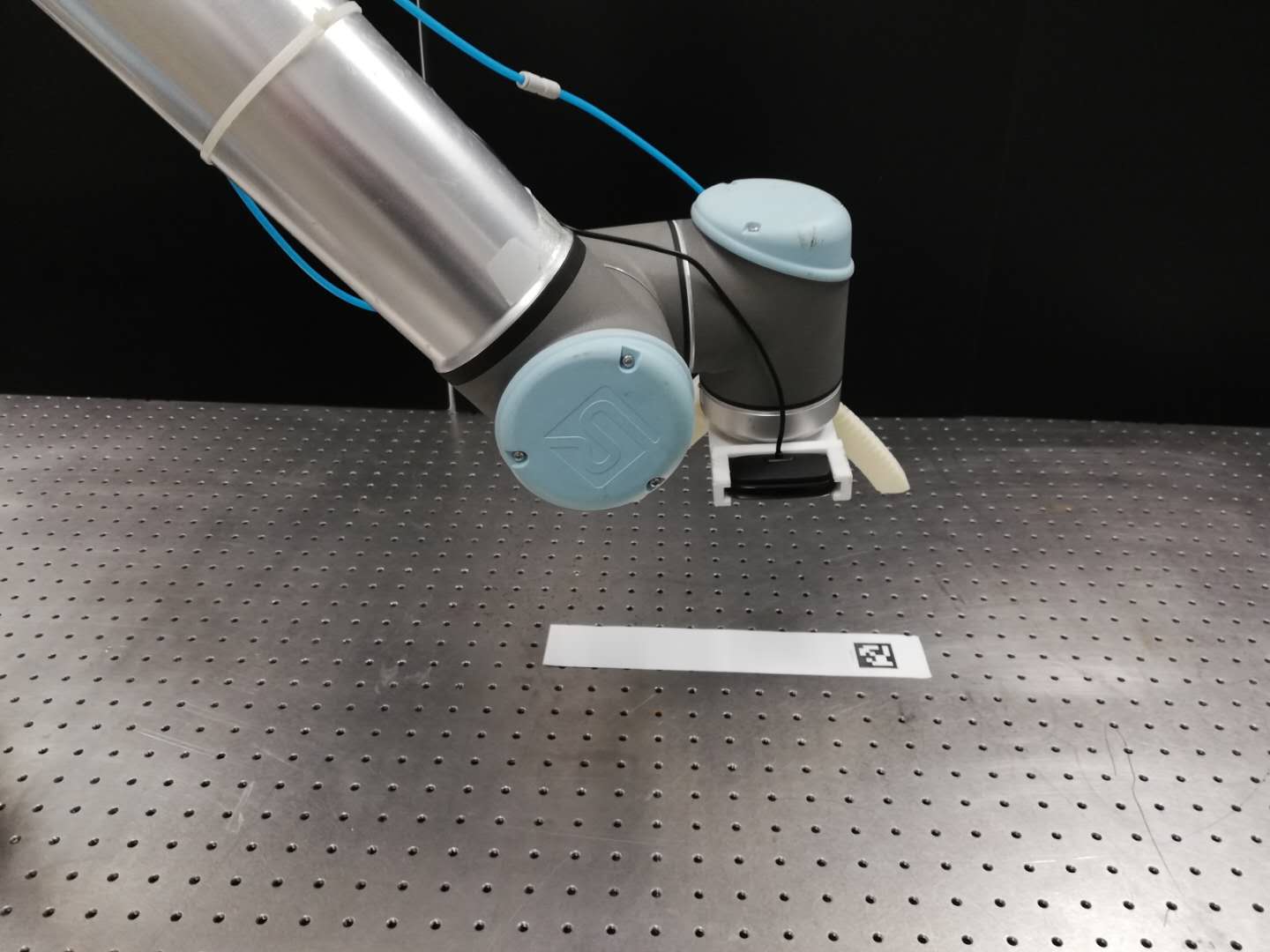}
\caption{}
\end{subfigure}

\caption{(a) Our two-fingered soft hand on the industrial robot arm UR10. (b) Hardware setup for robotic page-turning. Our gripper is  to  be  controlled  to  turn  the  paper  strip,  localized  autonomously through AprilTag. }
\label{fig:eehand}
\vspace{-0.2cm}
\end{figure}
%%%%%%%%%%%%%%%%%%%%%%%%%%%%%%%

This section presents the implementation of our flex-and-flip manipulation technique with a set of experiments.

\subsection{Our Two-Fingered Soft Hand}
\label{subsec:our_gripper}

Fig.~\ref{fig:eehand}a shows our soft hand to be used in our experiments. It has two underactuated, compliant fingers individually controlled by pneumatic actuation (SMC electro-pneumatic regulator ITV 0030-2S) and an RGB camera (Logitech webcam C525) on the palm. The fingers instantiate the idea of the ``fast pneu-net'' presented in \cite{Mosadegh_2014}. Basically, each finger incorporates a series of inflatable pneumatic chambers and is fabricated from elastomeric material, thermoplastic polyurethane, using a 3D printer, Ultimaker 2+. The fabrication process was guided by \cite{Yap_2016}.
\begin{figure}[b]
\centering         
\includegraphics[trim={1cm 0cm 2cm 2cm}, clip,width=0.45\textwidth]{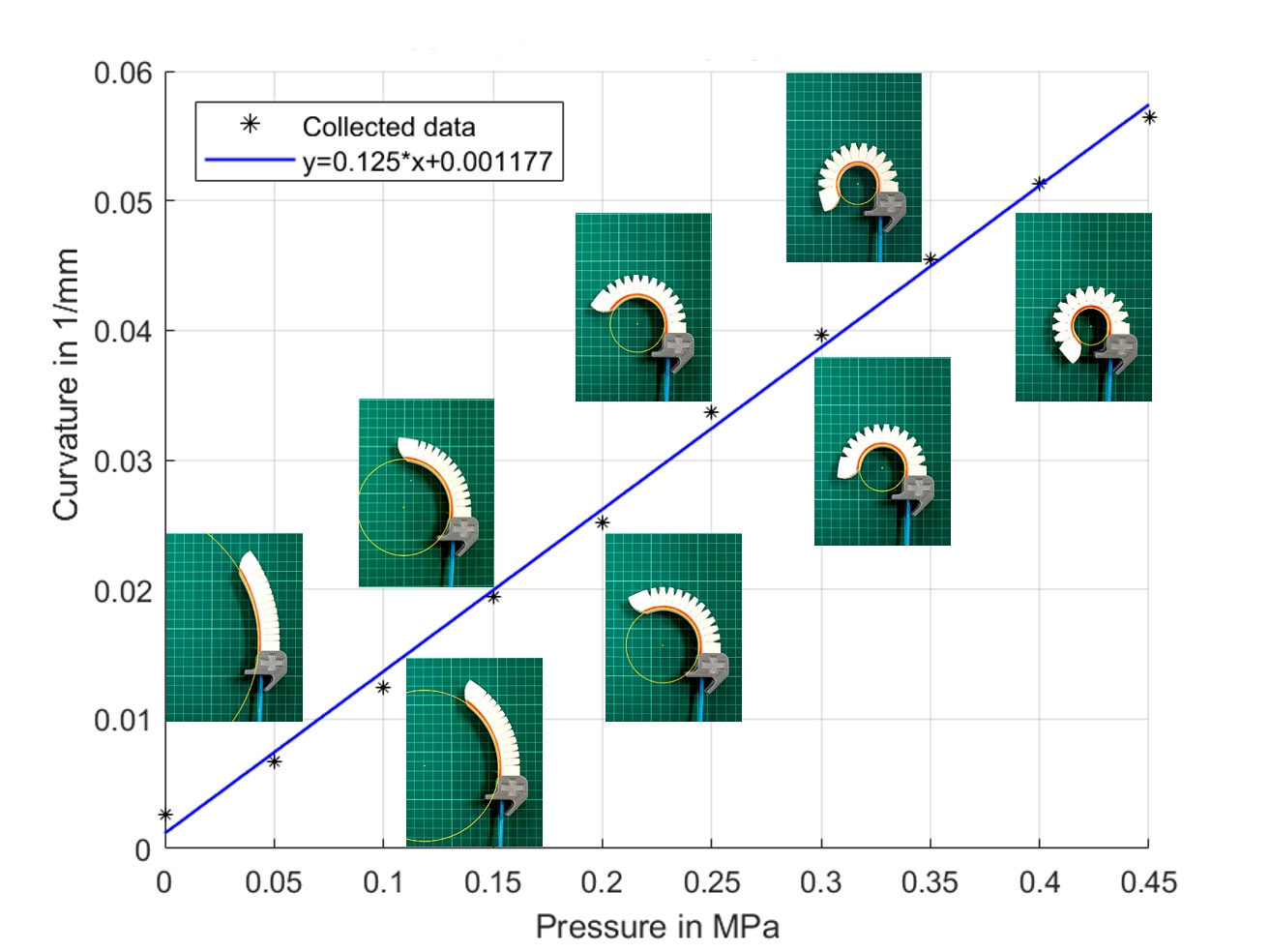}

\caption{Highly linear curvature-pressure relationship of our finger. The curvature of the finger was estimated through image processing using OpenCV (\url{http://opencv.org}).}
\label{fig:curvature_vs_pressure}
\vspace{-0.2cm}
\end{figure}
%%%%%%%%%%%%%%%
The overall shape of the fabricated finger can be represented as a circular arc that curls towards the palm. Its curvature is proportional to the internal pressure. The relationship between the curvature and the internal pressure of our finger is highly linear as can be seen in Fig.~\ref{fig:curvature_vs_pressure}. Angle between the fingers at the base is set for $90^\circ$ (Fig.~\ref{fig:eehand}a). The angle determines the size and shape of the pocket to accommodate the object during the flip operation, which can be critical to the outcome of the manipulation. Our experiments to be presented show that $90^\circ$ can be a feasible choice.

\subsection{Setup for Robotic Page-Turning Experiments}

Our flex-and-flip manipulation technique is tested on the task of page turning with our soft hand. Fig.~\ref{fig:eehand}b shows our hardware setup with the object, a strip of ISO A4 paper measuring 295mm by 50mm in size and 80 gsm (grams per square meter). The manipulation of page turning is supposed to happen on the plane formed by the two fingers and the strip, and thus perpendicular to the tabletop. First, the arm is controlled to move the hand to a desired initial configuration, an element of $SE(2)$ on the plane. Table~\ref{tab:tab1} illustrates the configuration with parameters $x$, $z$ (measured from the right tip of the object), and $\theta$, and their range to be tested. Second, fingers \#1 and \#2 (shown in Table~\ref{tab:tab1}) are pressurized to manipulate the page in an effort to obtain a pinch grasp as Fig.~\ref{fig:bifurcation}c.
%%%%%%%%%%%%%% Figure:1D turning model

\begin{table}
    \centering
    \begin{tabular}{l*{3}{c}c}
    \hline
    Parameters       & Min & Max & Step Size  \\
    \hline
    Horizontal Distance, $x$ (mm)     & 30 & 90 & 10  \\
    Vertical Distance, $z$ (mm)    & 116 & 135 & 1   \\
    Wrist Angle, $\theta$ (deg)    & 0 & 12 & 1   \\
    \end{tabular}
    \centering

    \includegraphics[trim={3cm 0cm 3cm 0cm}, clip, width=0.24\textwidth]{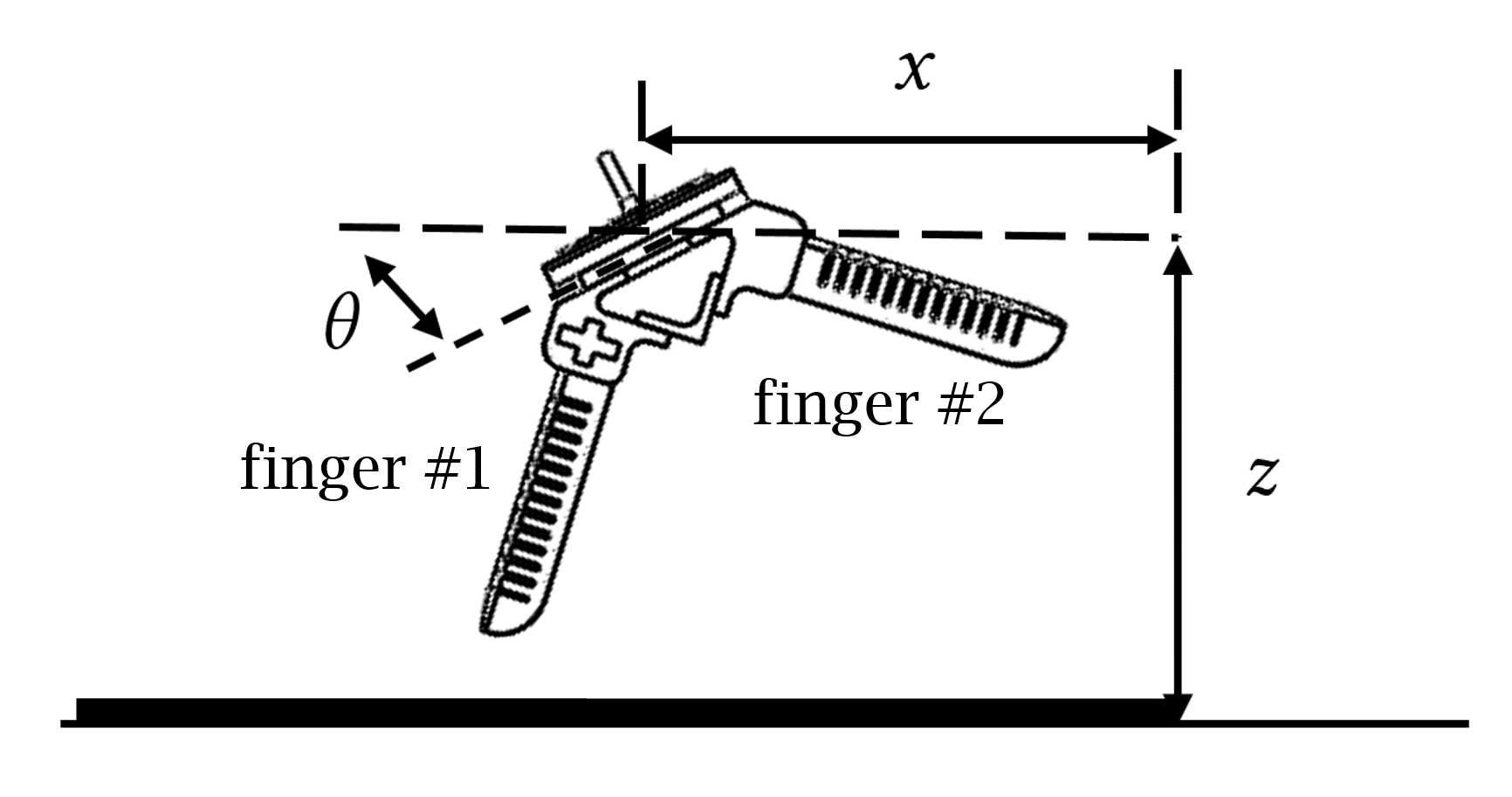}

    \setlength{\unitlength}{1cm}
    \begin{picture}(0,0)
    \put(-5,0.5){\footnotesize paper}
    \end{picture}
    \caption{Range of the initial configurations of the hand to be tested in our page-turning experiments.}
    \label{tab:tab1}
    \vspace{-0.2cm}
\end{table}
%%%%%%%%%%%%%%%

The experiments can be performed autonomously by running our software that incorporates the controllers for the arm and the pneumatic regulators of the fingers. AprilTag \cite{olson2011tags} is used for localizing the object; see the tag printed on the page in Fig.~\ref{fig:eehand}b. As will be shown by our experiments, the material properties of the finger can be sufficient for successful page turning: the finger is sufficiently compliant to be able to bounce off by the flexed paper (Fig.~\ref{fig:energy_plot}) and the friction at the fingertip is large enough to keep the contact for a sufficiently long time (Fig.~\ref{fig:cof}).

\subsection{Result}

\begin{figure}[h!]
\centering
\hfill
\includegraphics[trim={10cm 2cm 9cm 2cm},clip,width=0.5\textwidth]{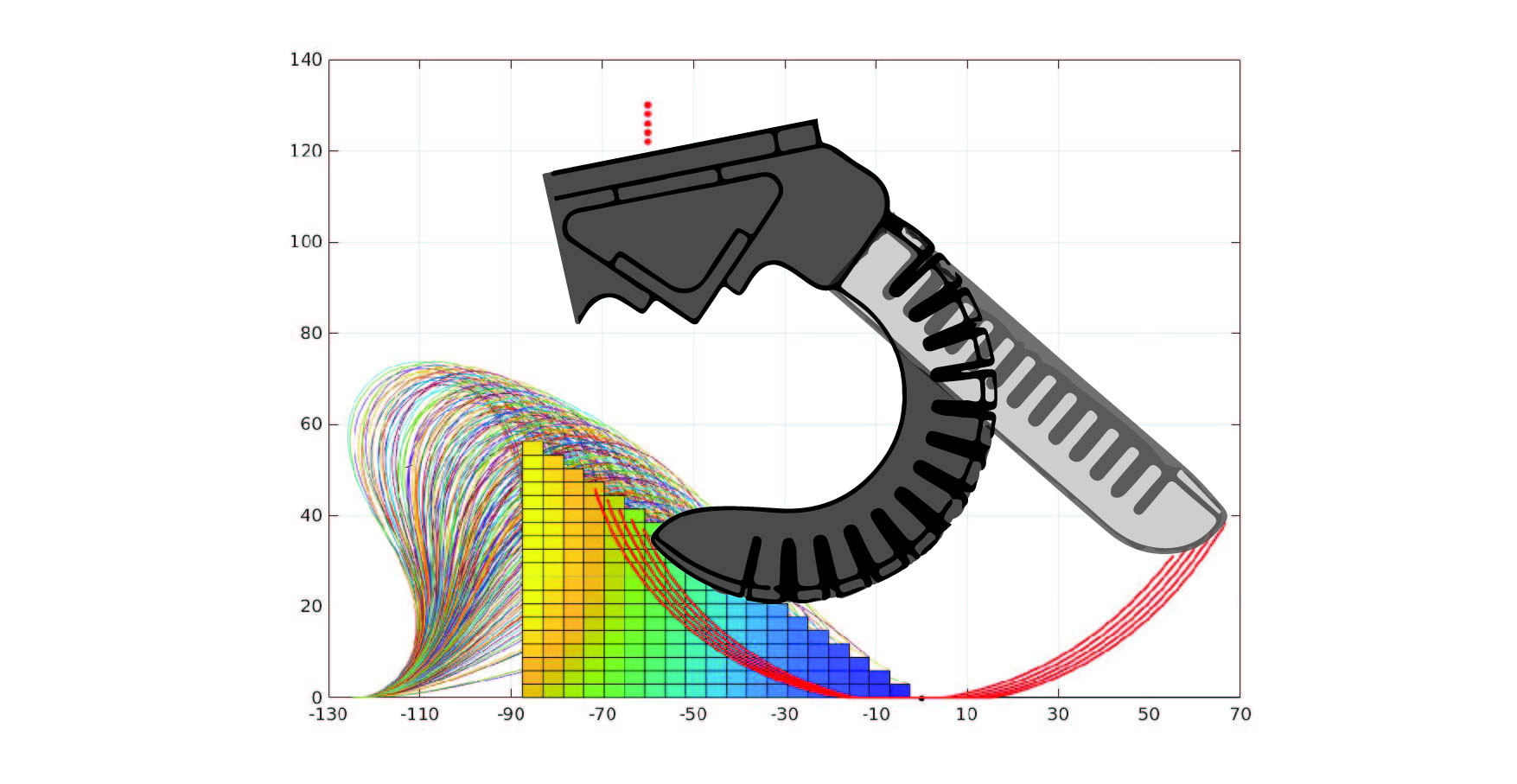}
\setlength{\unitlength}{1cm}
\begin{picture}(0,0)
\put(-1.4,0.1){\footnotesize Horizontal displacement (mm)}
\put(-4.3,1.7){\footnotesize \rotatebox{90}{Vertical displacement (mm)}}
\put(-0.70,5.7){\footnotesize config. 5: (60,130)}%$x=60$mm, $z=130$mm}
\put(-3.2,5.4){\footnotesize config. 1: (60,122)}%$x=60$mm, $z=122$mm}
\put(2.1,4.1){\footnotesize finger \#2}%$x=60$mm, $z=130$mm}
\put(2.2,1.4){\footnotesize path 5}
\put(3.1,1.0){\footnotesize path 1}
\end{picture}
\caption{As finger \#2 is pressurized, its tip nominally traces the paths shown as the red curves. The paths cannot penetrate the area below the horizontal axis (tabletop constraint); thus they feature a flat section on the axis. The five paths are based on the hand configurations $(x,z,\theta)$ determined by Eqs.~\ref{eqn:x} and~\ref{eqn:ztheta}. The red dots represent the $(x,z)$ coordinates of the configurations. Also shown is the dimensionalized energy plot derived from Fig.~\ref{fig:energy_plot}, where the length of the object is set at 125mm corresponding to our experiments.}
\label{fig:contact_simulation}
\vspace{-0.2cm}
\end{figure}
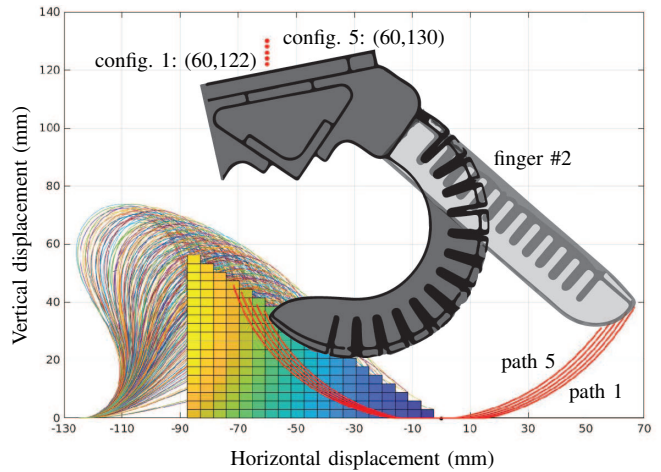

At each hand configuration $(x,z,\theta)$, the flex-and-flip manipulation is attempted by controlling the pneumatic regulators to apply pressure in the fingers: 0.15 MPa and 0.3 MPa for fingers \#1 (to fix the page) and \#2 (to flex the page), respectively. The system of the object and finger \#2 is then flexed in an open-loop manner. In total, $7$ (for $x$) $\times 20$ (for $z$) $\times 13$ (for $\theta$) $\times 5$ (repeated five times) $=9100$ attempts were conducted (see Fig.~\ref{fig:contact_simulation} for some example paths expected to be traced by the tip of finger \#2).

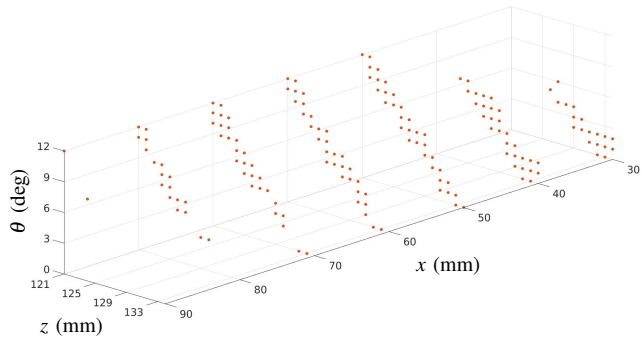
\begin{figure}[H]
\centering
\includegraphics[trim={3cm 2cm 3cm 2cm}, clip, width=0.48\textwidth]{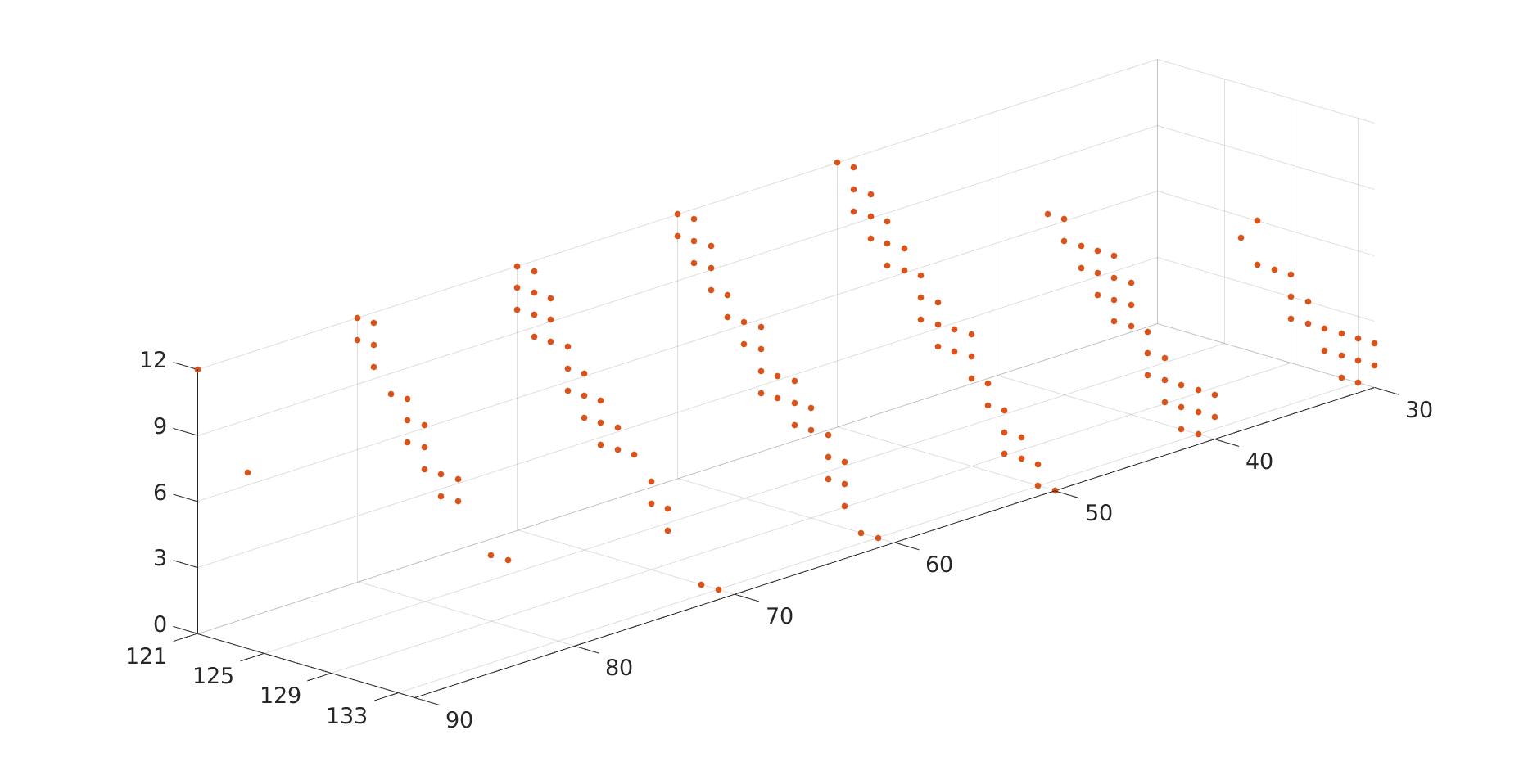}
\setlength{\unitlength}{1cm}
\begin{picture}(0,0)
\put(1.2,1.1){\footnotesize $x$ (mm)}
\put(-3.8,0.3){\footnotesize $z$ (mm)}
\put(-4.2,1.6){\footnotesize \rotatebox{90}{$\theta$ (deg)}}
\end{picture}
\caption{Hand configurations $(x,z,\theta)$ resulting in successful flex-and-flip manipulation.}
\label{fig:1D_result}
\vspace{-0.2cm}
\end{figure}

Fig.~\ref{fig:1D_result} shows the results of the experiments. The data points in the plot represent the set of the configurations $(x,z,\theta)$ whose success rate is equal to or greater than 0.6 (three or more successful flex-and-flip out of five trials). The clustered data points show that our flex-and-flip manipulation can be performed in a repeatable, robust manner. In failed trials, the fingertip did not interact with the paper in case $z$ was set too large, the fingertip was stuck on the tabletop in case $z$ was set too small, or $x$ (and thus $\delta$) was set too large for the tip of the object to be accommodated in the pocket formed by finger \#2 (recall our discussion in Sec.~\ref{subsec:bs}).

%%%%%%%%%%%%%%%       Figure    : 1D result

In Fig.~\ref{fig:1D_result}, it can be observed that (1) the relationship between $z$ and $\theta$ is modeled using an affine equation and (2) if $x$ is set too large/small, the range of $(z,\theta)$ becomes smaller. An optimal hand configuration $(x,z,\theta)$ for successful flex-and-flip manipulation can then be established based on these observations. First, select $x$ in the interval:
\begin{equation}
50\leq x~\textrm{(mm)}  \leq 70
\label{eqn:x}
\end{equation}

such that the feasible range of $z$ and $\theta$ can be maximized according to Fig.~\ref{fig:1D_result}. Second, select $z$ and $\theta$ that satisfy the affine equation obtained by the method of least squares:
\begin{equation}
\hat{\theta}~\textrm{(deg)} = -0.90z~\textrm{(mm)} + 120.5
\label{eqn:ztheta}
\end{equation}
in the range reported in the experiments. The nonzero size of $x$'s interval and the property of the obtained affine function---minimizing the sum of squared residuals---imply the robustness of the outcome of the manipulation performed with the selected parameters.

Fig.~\ref{fig:contact_simulation} confirms our discussion in Sec.~\ref{sec:theory} by showing what can happen in successful trials. It features some of the nominal paths to be traced by the tip of finger \#2, overlaid on the energy plot that also appeared in Fig.~\ref{fig:energy_plot}. The paths are obtained under the assumption that the finger flexes with uniform curvature (as can be seen in the insets of Fig.~\ref{fig:curvature_vs_pressure}). The tip will then move along a spiral curve. Here the configuration of the hand $(x,z,\theta)$ is set as discussed in Eqs.~\ref{eqn:x} and~\ref{eqn:ztheta}. As the fingertip moves along the path to the left, the flexural energy of the object will increase and there will be a point where the fingertip finally stops moving and bounces off from the object. If the fingertip is assumed to recoil along the path reversely, it can be seen that towards the end of the path the direction of the fingertip's motion may differ noticeably from the gradient of the flexure energy field. The difference can make flipping happen by guiding the tip of the page towards the pocket of finger \#2 (the area enclosed by the curled finger), as the fingertip (the object) moves with a relatively large downward (rightward) velocity. It can also be seen in Fig.~\ref{fig:contact_simulation} that the fingertip interacts with not only the object but also the ground where the path is shown flat on the horizontal axis. The successful results show that our compliant finger can appropriately address it. See also the video attachment, providing a slow-motion clip elaborating the discussion here.

%%%%%%%%%%%%%% Figure:contact simulation
%%%%%%%%%%%%%%%%

%%%%%%%%%%%%%%% Figure:  dynamic snap shot
\begin{figure*}[t]
\centering
\begin{subfigure}{15.5cm}
\centering
\includegraphics[width=16cm]{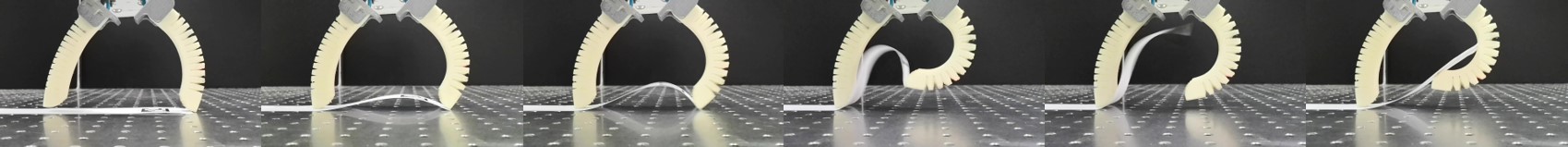}
\caption{Successful 1D (with a paper strip) flex-and-flip manipulation $(x=60\mathrm{mm},z=130\mathrm{mm},\theta=3^{\circ})$}
\label{fig:2D_12}
\end{subfigure}

\begin{subfigure}{15.5cm}
\centering
\includegraphics[width=16cm]{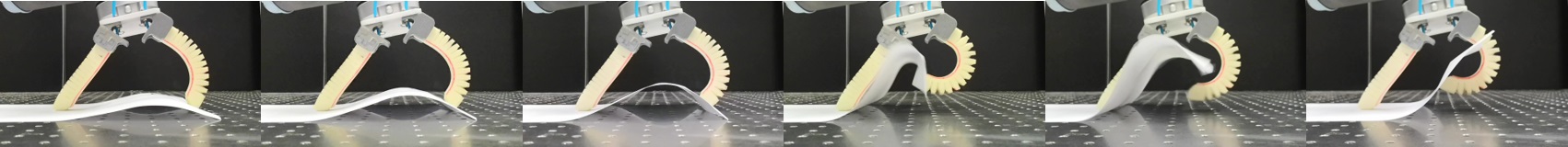}
\caption{Successful 2D (with a whole A4 paper) flex-and-flip manipulation $(x=50\mathrm{mm},z=120\mathrm{mm},\theta=12^{\circ})$}
\label{fig:2D_12}
\end{subfigure}

\begin{subfigure}{15.5cm}
\centering
\includegraphics[width=16cm]{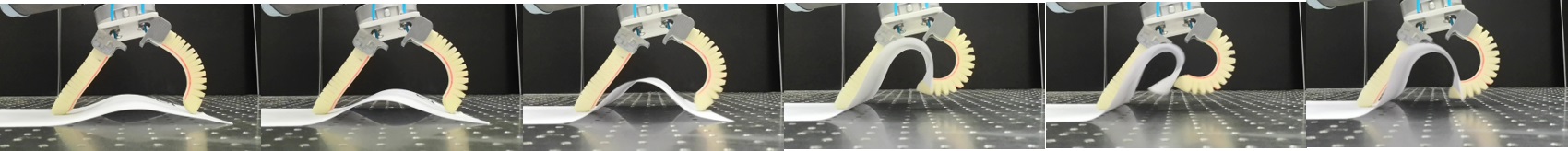}
\caption{Unsuccessful 2D flex-and-flip manipulation $(x=60\mathrm{mm},z=120\mathrm{mm},\theta=12^{\circ})$}
\label{fig:2D_12_unsuccess}
\end{subfigure}

\begin{subfigure}{15.5cm}
\centering
\includegraphics[width=16cm]{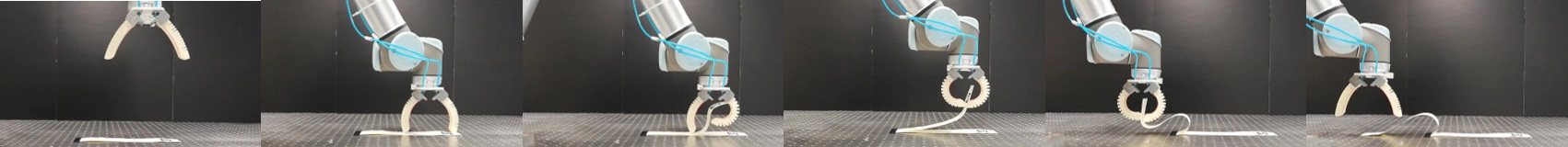}
\includegraphics[width=16cm]{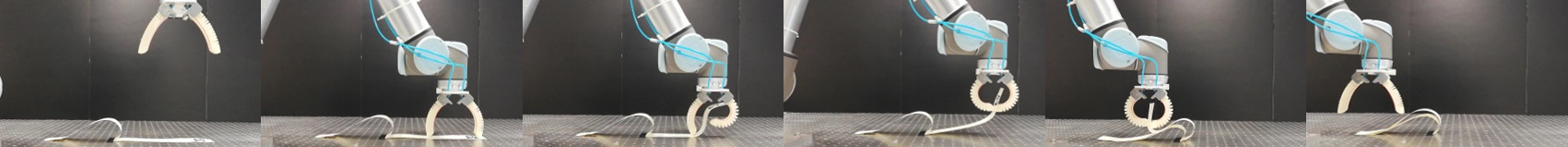}
\caption{Back-to-back page turning (left to right, top to bottom)}
\label{fig:2page_1}
\end{subfigure}

\caption{(a)-(c) Snapshots of flex-and-flip operations. (d) Successive page turning based on flex-and-flip.}

\label{fig:tuck-in snap shot}
\vspace{-0.4cm}
\end{figure*}

Fig.~\ref{fig:tuck-in snap shot}a shows successive snapshots of a successful flex-and-flip operation. In the second panel of the figure, finger \#2 (the one on the right) starts interacting with the paper strip (and also the tabletop) as it is pressurized. From the second to third panel, the finger and the ground ``fight against each other'' and thus the finger deforms accordingly in a quasistatic manner. As the finger separates from the ground, the potential energy accumulated during the interaction with the ground is converted into the finger's kinetic energy. The finger now moves dynamically until the moment shown in the fourth panel, in which the flex phase terminates and the flip phase begins. In the fifth and sixth panel, the finger recoils back downward as the tip of the paper strip moves rightward into the pocket area enclosed by the finger, as discussed in the previous paragraph. The flex-and-flip technique is also applied successfully to the whole A4 paper as shown in Fig.~\ref{fig:tuck-in snap shot}b, while the larger $x$ value (and thus larger $\delta$) can cause failure in Fig.~\ref{fig:tuck-in snap shot}c. Fig.~\ref{fig:tuck-in snap shot}d shows back-to-back complete page turning advanced from the flex-and-flip operation. See also the video attachment.

Finally, our flex-and-flip manipulation technique is also tested on other paper samples with different properties. The initial hand configurations are selected by Eqs.~\ref{eqn:x} and~\ref{eqn:ztheta} again. Generally, it can be seen that the success rate gets higher as the stiffness of the paper sample (that is, gsm) increases. 

\begin{table}
\small
\centering
\caption {Page turning experiments performed on a range of paper samples} \label{tab:tab2} 
\begin{tabular}{l*{2}{c}c}
\hline
Type of paper                & Successes/Trials & Success rate  \\
\hline
Writing Paper (70 gsm)      & 51/65   &78.5\% \\
Printing paper (80 gsm) &56/65 &86.2\% \\
Laid Paper (100 gsm)         & 58/65  & 89.2\% \\
Texture Paper (140 gsm)       & 62/65  & 95.4\% \\
Kraft Paper (160 gsm)     & 59/65 & 90.7\%  \\
Color Cardboard (200 gsm)     & 61/65  & 93.8\%  \\
\end{tabular}
\vspace{-0.4cm}
\end{table}
%%%%%%%%%%%%%%%%%%%%%%%%%%%%%%%%%%%%%
%%%%%%%%%%%%%%%%%%%%%%%%%%%%%%%%%%%%%%%%%%%%%%%%%%%%%%%%%%%%%%%%% CONCLUSION

 \section{Conclusion}
In this paper, we presented the technique of flex-and-flip manipulation for grasping flexible objects by leveraging the dynamics of object deformation. The technique is demonstrated through a set of experiments on the task of page turning. Possible directions for future work include enhancing robustness via tactile feedback to estimate friction forces and generalizing to the grasping of deformable 2D planar objects.

\bibliographystyle{IEEEtran} 
\bibliography{IEEEabrv,references}

\end{document}